\documentclass[conference]{IEEEtran}
\IEEEoverridecommandlockouts
\usepackage{cite}
\usepackage{amsmath,amssymb,amsfonts}
\usepackage{algorithmic}
\usepackage{graphicx}
\usepackage{textcomp}
\usepackage{xcolor}
\usepackage{hyperref}
\def\BibTeX{{\rm B\kern-.05em{\sc i\kern-.025em b}\kern-.08em
    T\kern-.1667em\lower.7ex\hbox{E}\kern-.125emX}}
\begin{document}

\title{ Towards Understanding the Link Between Modularity and Performance in Neural Networks for Reinforcement Learning 

%

}
\author{\IEEEauthorblockN{Humphrey Munn and Marcus Gallagher}
\IEEEauthorblockA{\textit{School of Information Technology and Electrical Engineering} \\
\textit{The University of Queensland}\\
Brisbane, Australia \\
h.munn@uq.edu.au, marcusg@uq.edu.au}
}

\maketitle

\begin{abstract}

Modularity has been widely studied as a mechanism to improve the capabilities of neural networks through various techniques such as hand-crafted modular architectures and automatic approaches. While these methods have sometimes shown improvements towards generalisation ability, robustness, and efficiency, the mechanisms that enable modularity to give performance advantages are unclear. In this paper, we investigate this issue and find that the amount of network modularity for optimal performance is likely entangled in complex relationships between many other features of the network and problem environment. Therefore, direct optimisation or arbitrary designation of a suitable amount of modularity in neural networks may not be beneficial. We used a classic neuroevolutionary algorithm which enables rich, automatic optimisation and exploration of neural network architectures and weights with varying levels of modularity. The structural modularity and performance of networks generated by the NeuroEvolution of Augmenting Topologies algorithm was assessed on three reinforcement learning tasks, with and without an additional modularity objective. The results of the quality-diversity optimisation algorithm, MAP-Elites, suggest intricate conditional relationships between modularity, performance, and other predefined network features. 

\end{abstract}


\section{Introduction}

Modularity plays a crucial role in the structure, function, and scalability of biological neural networks, particularly in the human brain \cite{b1,b2,b3}. In artificial neural networks, modularity has been a continuing topic of interest in areas such as neural architecture search (NAS) neuroevolution, as it is one of the network properties they aim to optimise \cite{b4,b5}. Although some methodologies have been proposed to leverage modularity to enhance neural network performance, the results have been surprisingly sparse. Modularity is a key characteristic in various complex systems such as road networks \cite{b6}, organizations \cite{b7}, and social networks \cite{b8}, as it promotes diversity, leverages similarities, and simplifies complexity \cite{b9}. Therefore, these benefits may have potential to improve the ability of artificial neural networks. While some general structural constraints have been a part of artificial neural networks and deep learning (e.g., convolutional layers), a large amount of work utilises non-modular neural networks (e.g., fully-connected feedforward networks). Our current understanding of the utility and importance of modularity in artificial neural networks is limited.

In this paper, we investigated the role of modularity in networks generated via neuroevolutionary techniques for reinforcement learning problems. Specifically, we use a measurement of modularity based on the graph representation of the neural network, so modules are intrinsic to the network architectural properties and are not manually defined. Our neural networks are not constrained to the standard layered architecture, thus promoting rich exploration of network configurations. 
Our results show significant dependencies between performance, modularity, and many other network properties that were tested. We evolved a MAP-Elites grid of neural network solutions which displayed the performance landscape over modularity and the other defined network features. We found that adding an additional weighted reward for modularity to the learning algorithm increased the modularity of solutions in proportion to the reward amount, but did not improve performance. However, in experiments where network optimisation using the NEAT algorithm was performed, modularity and performance both increased over learning epochs (generations), suggesting that modularity was required as it had been selected for in the evolutionary optimisation process. These findings demonstrate that modularity can improve the performance of neural networks, but its design and amount are difficult to determine, and selecting modularity naively in an architecture is unlikely to be beneficial.  


The rest of the paper is organised as follows. Section~\ref{sec:bg} discusses background material, including modularity calculations and explanations of the algorithms used. Section~\ref{sec:method} describes our experimental methodology. We present our results in Section~\ref{sec:results}, followed by a discussion in Section~\ref{sec:discussion} and conclusions in Section~\ref{sec:conclusion}. 

The source code for the experiments in the paper is available at \url{https://github.com/humphreymunn/ModularityNEAT/}.

\section{Background}
\label{sec:bg}
\subsection{Modularity}

Modularity is an ill-defined term, whose meaning is often context-specific. Generally, it is the property of a system which displays varying density of connectivity between system parts, with high modularity implying dense clusters or communities that are sparsely connected to others. In the context of neural networks, this connectivity may be defined as the connections between neurons. Standard artificial neural network architectures are not designed with a high degree of modularity, however large, deep networks present major challenges in terms of resources and training. Large leaps in neural network ability have been afforded by innovative changes to the structure of neural networks, such as convolutional layers \cite{b10}, recurrent connections, and transformers \cite{b11}. The structural advantage of modularity in neural networks has shown some promise, however there is currently a limited understanding of when and why it works most effectively. This paper will aim to shed light on this issue. 

The significant modularity exhibited in complex systems seen in nature and the modern world, including the neural network of the brain, suggest that a greater investigation of this property could be beneficial for future neural network designs. 

There are many studies which have shown advantages in inducing modularity effectively. In \cite{b12}, state-of-the-art generalisation performance is achieved on visual question answering datasets by designing a modular neural network architecture with modules that were trained explicitly for different sub-functions. Furthermore, \cite{b13} finds that high degrees of modularity can mitigate catastrophic forgetting and links this with biological plausibility. A modular ANN variant is proposed in \cite{b14}, aimed at hierarchical representation of the objective function. They find that it produces interpretable, tree-like functionality, which enables function composition and high performance on complex tasks. Routing networks have been proposed for effective compositionality of different tasks through induced modularity, but it was found in \cite{b15} that structural bias was necessary to achieve maximum performance. It has been hypothesised that the modular routing network architecture will lead to the next revolution in AI \cite{b16}.

To the best of our knowledge, none of the current literature attempts to explain the difficulties of neural network modularity optimisation. Yet, there are many demonstrations of it succeeding and not succeeding through different schemes. The findings from \cite{b17} offer a partial explanation, suggesting that functional specialisation is tied to structural modularity even without explicitly inducing this relationship. This may explain why different problems have such distinct responses to modular neural network solutions, often requiring entirely different architectures or training schemes to perform well between problems, and often do not benefit from the tested amounts of modularity at all.

\subsection{Measuring Modularity}
The concept of neural network modularity is developed using ideas from graph theory and the community detection \cite{b8, b17, b18}. As fully-connected neural networks are analogous to directed acyclic graphs, the same application of modularity detection and measurement can be applied. In this case, neurons are represented as nodes and connections between neurons as edges. The degree of modularity of a network can be measured by its Q-Score \cite{b8, b19}. Q-Score is obtained by maximising the Q function by partitioning the set of all network neurons into disjoint sets (i.e modules) that have high inter-connectivity but low connectivity between modules. Q-score is defined as:
\begin{equation}
Q = \sum_{s=1}^{K}\left[\frac{l_s}{L} - \left(\frac{d_s{^2}}{2L}\right)\right],
\end{equation}
where $K$ is the number of modules, $L$ is the number of edges, $l_s$ is the number of edges within module s, and $d_s$ is the sum of the degree of each graph nodes in s. The $Q$ function is bounded by $[0,1]$, with 0 corresponding to no modularity and 1 being maximum modularity. Direct calculation of Q-score is an NP-Hard problem \cite{b20}, so a close approximation of the optimal Q-Score is computed using Newman's method \cite{b8}. 
Intuitively, maximising the Q-Score will find optimal groupings of neurons into modules with high connectivity, and this maximal score determines the network modularity.



\subsection{NeuroEvolution of Augmenting Topologies (NEAT)}

Neuroevolution enables automatic optimisation of both network topology and weights, which makes it more suited for an investigation into the effects of topology on performance compared to fixed-topology methods such as backpropagation. The networks created by NEAT \cite{b21}, a neuroevolutionary algorithm, are feed-forward networks however they are more general in that they need not conform to a layered architecture. Therefore, it is reasonable to assume that properties and relationships that exist in the resulting networks could extend to networks from other learning algorithms. While networks created with NEAT are often small (typically hundreds of neurons or less), they have been applied to a number of complex problems with success. NEAT appropriates several concepts from natural evolution to solve neuroevolutionary challenges, such as the competing conventions problem for recombination of parent networks, protection of structural innovation via speciation, and gradual complexification of network structure for efficient network selection. With widespread usage in the research community and diverse search of network architectures, NEAT was a suitable contender for modularity experimentation. There are numerous extensions and variants of NEAT, including HyperNEAT \cite{b22} and Coevolutionary NEAT \cite{b23} that typically exceed the performance of the canonical implementation. However, the standard NEAT algorithm was enough for our purposes. 





\subsection{Reinforcement Learning Environments}

In our experiments, we have used three benchmark reinforcement learning tasks, as they require complex control and are well-suited to the NEAT algorithm. These were Bipedal Walker, Continuous Lunar Lander, and Acrobot \cite{b24}. 

\begin{figure}[!h]
\label{fig:openaiproblems}
\centering
\fbox{\includegraphics[width=1.0\linewidth]{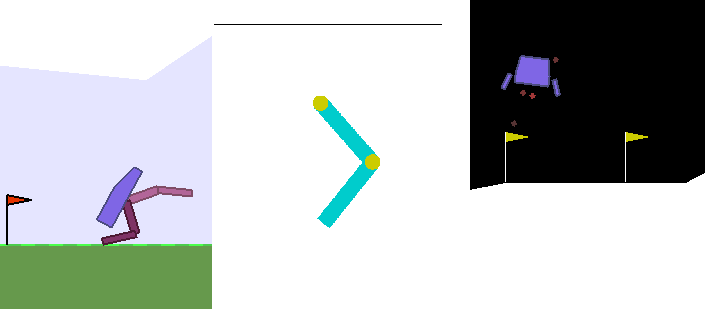}}
\caption{OpenAI test environments. Left: BipedalWalker, Middle: Acrobot, Right: ContinuousLunarLander.}
\end{figure}

The bipedal walker task requires the agent to walk from left to right in the environment (Fig.1, left). Agent reward is simply the distance travelled in the time limit, with a penalty for falling over. The state space consists of 24 values: 10 lidar readings, each joint angle and speed, leg ground contacts, $x$ and $y$ velocity, hull angle and hull angular velocity. The action space, $A$ or output is the torque to be applied to each of the two hip joints and two knee joints. That is, $A \in [-1,1]^4$.

The Acrobot task requires the agent to control an actuated joint to swing a two-link pendulum above a specified height (Fig.1, middle). The reward is simply how fast the agent can solve the task. The state space consists of the sine and cosine for each joint angle, and the angular velocity for each joint. The action at each time step is the torque to be applied on the joint, where the action space is discrete. That is, $A \in \{-1, 0, 1\}$.

The continuous lunar lander task requires the agent to navigate a lunar lander to the landing pad at coordinates (0,0) in a simulated physics environment (Fig.1, right). Agent rewards depend on how close the landing is to (0,0), how much throttle is used (to maximise fuel efficiency), and whether the agent lands safely. The state space is given by the Lander's coordinates (x,y), velocities (x,y), angle and angular velocity, and left leg and right leg contact with the ground (True/False). The action space is two values: 1. the main engine throttle, and 2. the left/right engine throttle (left is in [-1,-0.5], right is in [0.5, 1]), where $A \in [-1,1]^2$.

\subsection{MAP-Elites}

MAP-Elites \cite{b25} is a recent idea in evolutionary algorithms that creates a map or grid of high-performing solutions, where the dimensions of the grid are important features of the problem decided by the user. In this paper for example, network modularity is one of these important features. Each grid square represents the highest fitness individual whose features are within the ranges defined by the grid cell position, and the colour is the individual's fitness. The algorithm generates an initial random population, which is added to the grid. Then at each generation, a set number of individuals are procreated and mutated to create offspring, which are then added to the grid. If two individuals overlap on the grid, the highest fitness individual is kept. This strategy has been shown to often outperform direct optimization of performance, as it typically explores more of the search space, biased by important features \cite{b25}. Furthermore, it provides information about the fitness landscape of individuals, revealing important features of the applied problem or genotype/phenotype mapping. By illuminating the landscape over modularity and other predefined features of either the resulting networks or behaviours from such networks, the entangled nature of modularity could be clearly visualised. 


\section{Methods}
\label{sec:method}

The modularity and fitness of solutions generated by the NEAT algorithm were measured on the three benchmark problems mentioned earlier. 20 runs of the learning process were performed for each problem. The NEAT hyperparameters used were the same as the neat-openai-gym GitHub repository (Rojas, 2017) as these were found to work well and did not need further tuning. However, population size was modified for faster convergence, with all problems using a size of 400 individuals. Averages over 5, 10 and 20-episodes were used during training for BipedalWalker, ContinuousLunarLander and Acrobot respectively to minimise noise when measuring fitness. The default feed-forward network configuration was used for the BipedalWalker and ContinuousLunarLander tasks, whereas Acrobot used a recurrent NEAT network. The best network at each generation was taken to generate the plots in Figure 2. 

In order to influence the modularity of the networks, an additional modularity reward was then added to the fitness function and assessed on the BipedalWalker task with varying levels of reward significance. This reward was scaled with the current fitness to bias increasingly modular networks as the population starts converging. An importance metric was defined, which was the maximum percentage of the current fitness that can be added to the fitness function. The new fitness function was defined as, 
\begin{equation}
F = R + Q \cdot I \cdot (min\{max\{R, a\}, b\} + a)
\end{equation}

where F = fitness, R = reward, Q = Q-Score, I = Q-Importance hyper-parameter, (a,b) = lower and upper fitness bounds. This results in a fitness function where an increase in environment reward will increase modularity desirability and visa-versa. 

Finally, the MAP-Elites algorithm was run on the BipedalWalker task, as this task was the most interesting due to its challenging nature. Various different features were tested along with modularity, with fitness being the average episodic reward achieved by the network. The grid was of size 20x20 to reduce the computational cost of a large grid. At each generation, 12 offspring were created with a mutation probability of 0.75 and crossover probability of 0.3 by random selection of solutions stored in the grid. These individuals were NEAT networks configured with the same hyperparameters as in the initial experiments. Changing the probabilities of mutation and crossover in preliminary experiments had negligible effects on the map. The 12 offspring were tested in parallel at each generation. This was run for 30000 generations, as we observed that the grids appeared to have mostly converged after this time.

\subsection{MAP-Elites Feature Descriptors}
The first feature tested was the "Deviation from Uniform Torque Output Distribution". For reinforcement learning tasks, the amount that each network output node is activated strongly impacts its behaviour. For the bipedal walking task, the output neurons correspond to the torque of each joint. So, the uniformity in the amount each joint is used is a useful descriptor of the walker's gait. For example, a local optima to the bipedal walking task could involve only applying torque to one knee joint. A simple method of approximating the deviation from a uniform distribution of the output neuron activations is used to obtain a network feature in the range [0,1]. The range rule of thumb is a common approximation to the sample standard deviation, such that $S \approx \frac{b - a}{4}$, where $b - a$ is the sample range \cite{b26}. An array of the total absolute torque applied to each joint over the episode time is stored, $X = [ x_1, x_2, ..., x_n]$, where $n$ = number of output neurons. The average over all episodes is taken. As the minimum range is 0, and maximum is $max\{X\}$, the normalized approximate deviation is then,
\begin{equation}
D = \frac{max\{X\} - min\{X\}}{max\{X\}}, D \in [0,1]
\end{equation}

The average contact time for each bipedal walker leg was also tested, see Fig. 4 (top). This was taken as an average of both legs, also averaged over episodes. This is another behavioural descriptor, that describes an important characteristic of the agent's gait. As the averages were taken from the contact point observations, the feature is in the range of [0,1] where 1 means constant contact and 0 is no contact. 

The final two features were the average x velocity of the walker, and the number of neurons in the network. To obtain a [0,1] scale for the number of neurons, a bound was put on the number of hidden neurons from 0 to 30, and then this was normalized into the stated range. 

Two ContinuousLunarLander features were also tested; the number of neurons and the percentage of the main throttle over the total throttle amount. These are briefly discussed in Section~\ref{sec:results}. 

\section{Results}
\label{sec:results}
\subsection{Characterising modularity and performance on standard NEAT optimisation}

\begin{figure}[!ht]
\centering
    \includegraphics[width=1\linewidth]{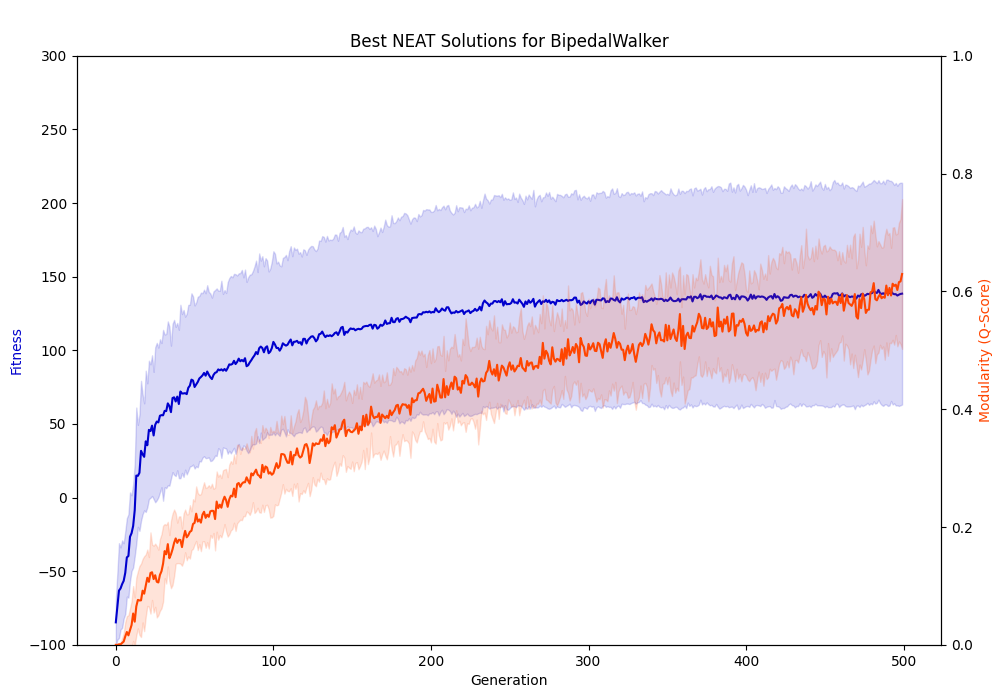}
    \includegraphics[width=1\linewidth]{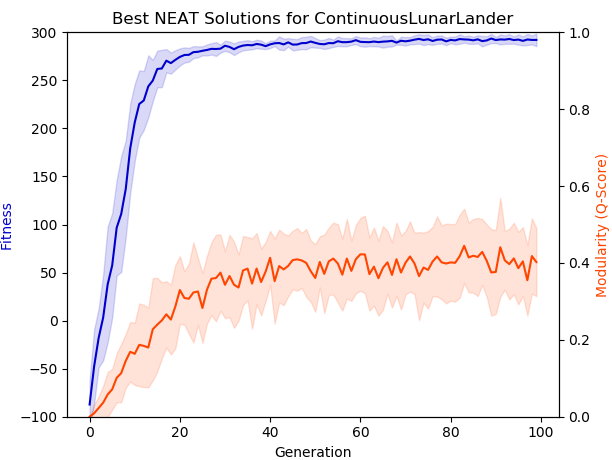}
    \includegraphics[width=1\linewidth]{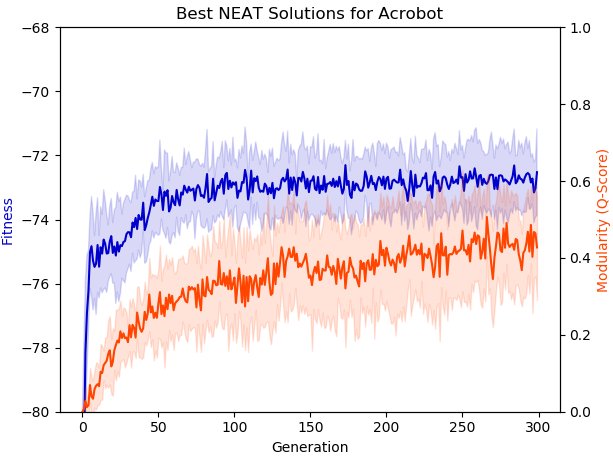}
\caption{Average fitness (reward) and modularity (q-score) over generations of the standard NEAT learning algorithm. Each test is 20 runs, with 1 standard deviation confidence intervals.}
\end{figure}

The modularity and fitness for the three reinforcement learning tasks using NEAT are displayed in Figure 2. In all three tasks, modularity increased over generations in the fittest individuals. The rate of increase and convergence differed between problems significantly. Both BipedalWalker and Acrobot experience a convergence of fitness but the modularity continues to increase beyond this convergence. Modularity appears to have converged in ContinuousLunarLander and Acrobot at the end of training but not for the BipedalWalker. Interestingly, the fitness of the BipedalWalker follows a bimodal distribution over runs at each generation suggesting it often gets stuck in locally optimal solutions.

\subsection{Adding modularity reward to the fitness}
Figure 3 (top) shows the best fitness for importance values (Equation 3) \{0, 0.1, 0.125, 0.15, 0.175, 0.2\}, averaged over 20 runs. There does not appear to be a linear relationship between a higher reward for modularity and higher performance. While some particular values of $I$ (0.175, 0.125) had higher average performance, this did not appear to be statistically significant. Figure 3 (bottom) shows the differences in how modularity evolves with these different reward amounts. In this plot, there is a clear linear relationship between increasing the reward for modularity and a higher level modularity. These results indicate that the reward function was effective at encouraging modularity in the solutions.

\subsection{MAP-Elites performance landscape}
While it is clear that modularity exists and (to some extent) can be influenced in the networks found by NEAT, the nature of the relationship between modularity, problem features and fitness seems to be more complex. Figure 4 clearly shows the level of modularity with the other respective features has an effect on the performance of the networks. In Figure 4  (bottom), two hot spots of high performing networks are visible within the ranges $ Q \in [0.25,0.5], D \in [0.55, 0.7]$. Modularity has reasonably performing solutions for all values, however $D$ (see Equation 2) is consistently poor past 0.8. This should correspond to the networks predominantly using a single output action out of the 4 available, i.e applying torque to only one of the joints. There are 2 main regions that are above 200 fitness, which correspond to different walking strategies. Figure 4 (top) also shows a high performing area in the grid within a range of modularity and the percentage airtime feature. All of the grids  show areas of high performance dependent on both grid features.

\begin{figure}[!htpb]
\centering
\includegraphics[width=0.89\linewidth]{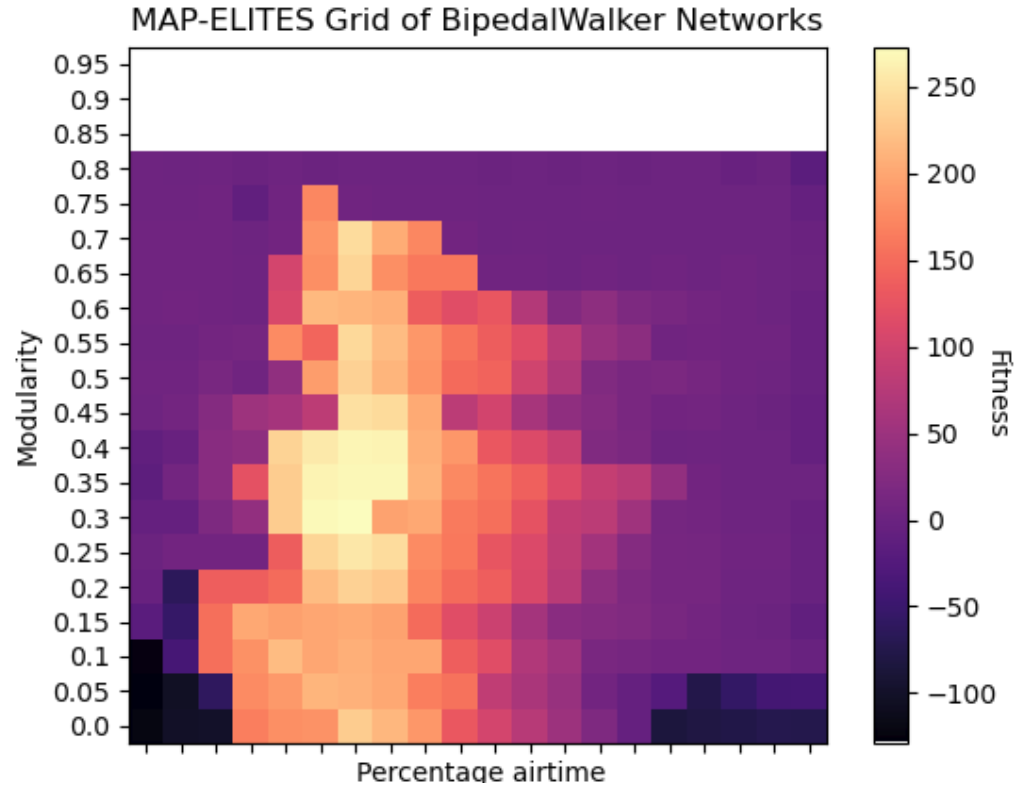}
\includegraphics[width=0.88\linewidth]{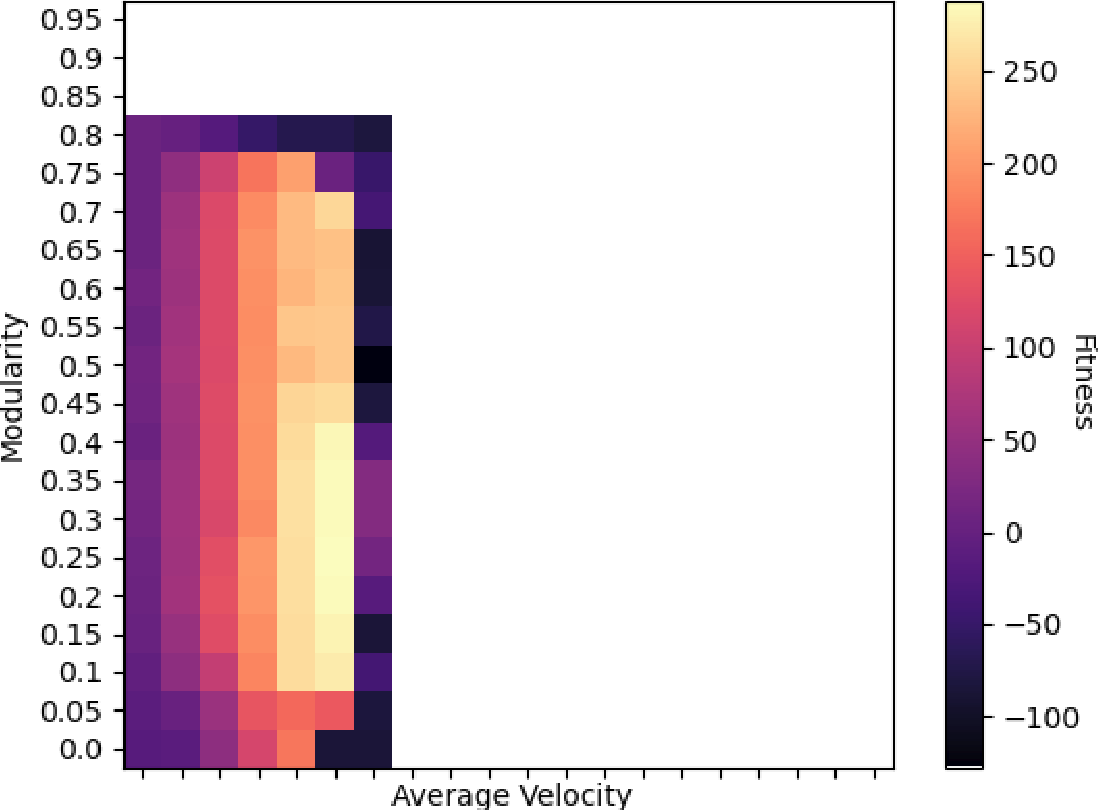}
\includegraphics[width=0.89\linewidth]{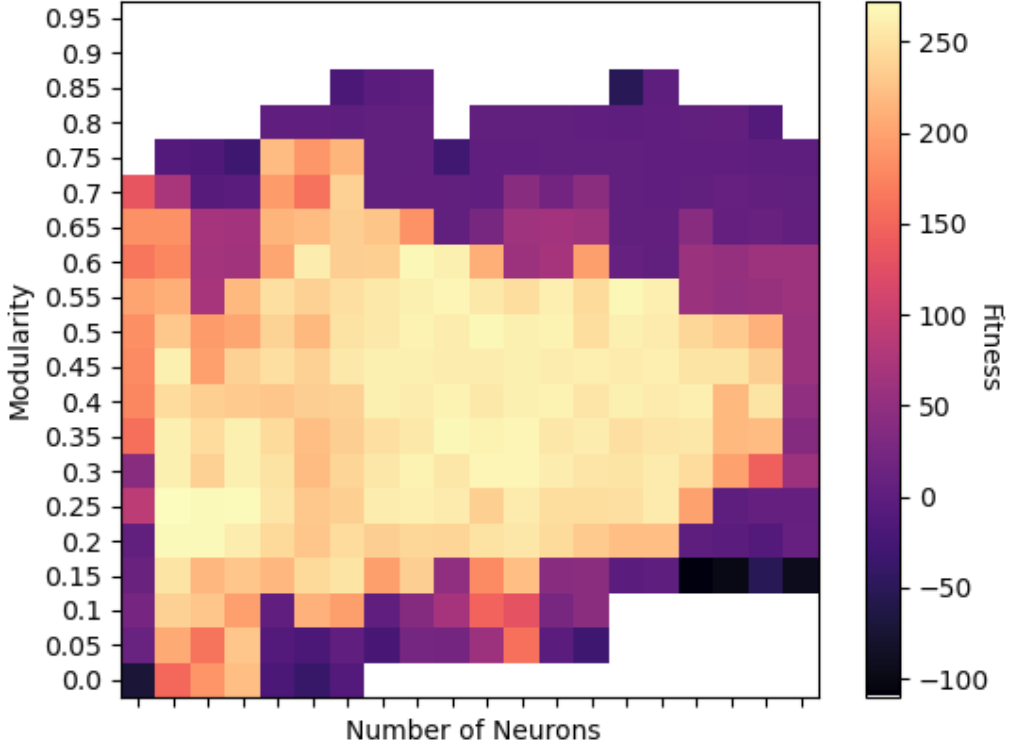}
\includegraphics[width=0.84\linewidth]{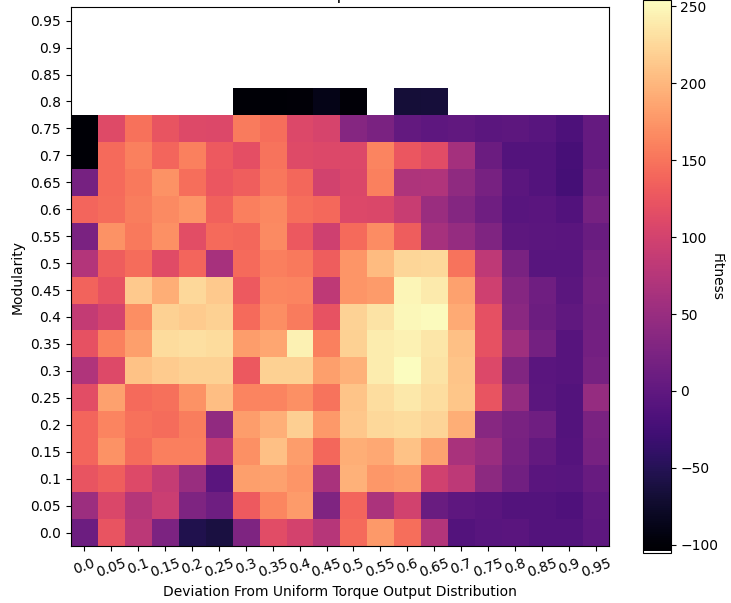}
\caption{MAP-Elites grid for BipedalWalker task showing the performance landscape over network modularity and other network features.}
\end{figure}

The two features of the ContinuousLunarLander environment that were tested, number of neurons and percentage of main throttle to total throttle, had limited to no entanglement between the respective features and modularity. This may be due to the environment being much easier to solve and significantly less complex than the bipedal walker (see Figure 2 (middle)).

All MAP-Elites experiments outperformed even the best run of the standard NEAT optimisation for the BipedalWalker. This was according to the episodic average of reward achieved by each network. Standard NEAT achieves a maximum fitness of 248 across all runs. The deviation from uniform torque output distribution experiment achieved a fitness of 254. The percentage airtime experiment achieved a fitness of 273. The number of neurons experiment achieved a fitness of 269. Finally, the average velocity experiment achieved a fitness of 288. This is close to solving the BipedalWalker, which requires an episodic average fitness of 300 to solve. 

\section{Discussion}
\label{sec:discussion}
The results of the first experiment showed that modularity provided a selective advantage to the evolution of the neural network controllers. This was evident by the upwards trend of modularity over training time in all three environments. Notably, the modularity of the highest-performing networks varied between environments. This is not surprising at it was shown previously that the optimal quantity of modularity tends to vary across problem environments \cite{b27, b28}. Some other works such as \cite{b29} also suggest that the network or learning algorithm itself influences the optimal level of modularity. This is further reinforced by our findings, as the Q-Score ranges of greatest performance vary between the direct training and MAP-Elites experiments. The direct training experiments show that modularity provides a significant performance advantage in these scenarios. 

The next experiment measured the effect of adding an extra reward for modularity onto the evolutionary process, to varying degrees. The degree of the importance hyper-parameter (see Equation 2) determined the scale of modularity rewards received. Figure 3 shows that the additional reward succeeded in creating more modular networks faster, but did not clearly impact the algorithm performance. This reinforces the initial hypothesis that taking a direct path optimising modularity would not provide performance benefits. While modularity is a common property of the best networks, balancing its function and performance can be complex and may require a specific design. This finding is reinforced in \cite{b30} where Q-Score was used as an guiding objective (Q-Mod) and this reduced performance on the baseline. However, when the diversity of modular decompositions of networks was used as an objective (ModDiv), this improved performance. This benefit may be explained by the complex modularity-performance relationship presented in this paper, as optimising the diversity of solutions explores more of this complex space.

Finally, the MAP-Elites experiments demonstrated that the way modularity impacted network performance was dependent on other network variables like their behaviour. Figure 4 shows four examples of this on the BipedalWalker task, where the feature axes were dependent in all cases. The reasons for the entanglement of modularity with these other features remains unclear. However, the identification of this property explains why modularity appears beneficial but often fails to improve networks when researchers design modular architectures or try to learn them. Specifically in the airtime, velocity and uniform torque map-elites grids, the areas of high performance in the grid are relatively small. The two lunar lander experiments (mentioned in Section~\ref{sec:results}) did not show such dependency along map-elites axes, however that may be due to the simplicity of the problem compared to the bipedal walker. Another interesting observation is that the number of neurons map-elites plot showed that the optimal modularity range decreased as the networks got bigger. This is congruent to the popular idea that modularity is a way of managing complexity in neural networks \cite{b31,b32}. 

The efficiency of modular neural networks is evident in the high performance of MAP-Elites solutions. The solutions found were much smaller than the state-of-the-art reinforcement learning algorithms and architectures typically required to achieve these rewards, such as TD3 \cite{b33} and PPO \cite{b34}. These generally require hundreds of hidden neurons to solve the bipedal walker task, whereas the MAP-Elites NEAT solutions come close to solving the problem using as little as 6 hidden neurons. MAP-Elites has already been shown to improve performance in the BipedalWalker task when useful feature descriptors are chosen, such as in \cite{b35} which evolves a grid of A3C networks. As modularity was an important determinant for performance, it should therefore be considered for future neural network research for efficiency. It may be that the lack of improvement from a modular architecture is a misleading factor, due to the complicated performance landscape of modularity.

\section{Conclusion}
\label{sec:conclusion}
Our study shows that modularity is inextricably linked with the performance of neural networks, although their relationship is nuanced. The complex interdependence between modularity and other network features makes it unlikely that simple selection of modular network designs or blind optimization of modularity will be effective. Three experiments demonstrated this: observing modularity emerging from evolutionary learning, demonstrating no performance improvement from an added modularity objective, and mapping the conditional effect of modularity and other features on performance using MAP-Elites. Furthermore, optimizing modularity while considering other network features significantly improved performance. Our findings suggest that modularity may be necessary for efficient network architectures but presents a challenging optimization problem. This view is reinforced by the significant levels of modularity seen in the human brain and other complex systems. Limitations include generalizing findings to backpropagation-trained networks and the need for further investigation in complex domains and the cause of entanglement. Future studies could explore complex network structures such as convolutional layers \cite{b17}, use larger networks, and incorporate network weights into the modularity calculations by using community detection algorithms such as \cite{b36}. 



\vspace{12pt}
\end{document}